\documentclass[journal]{IEEEtran}
\ifCLASSINFOpdf
\else
\fi
\usepackage{amssymb}
\usepackage{latexsym}
\usepackage{amsmath}

\usepackage{algorithm}
\usepackage{algorithmic}

\usepackage{url}
\usepackage{graphicx}
\usepackage{subcaption}

\begin{document}
%
\title{Fast Graph-Cut Based Optimization for Practical Dense Deformable Registration of Volume Images}
%
%
%


\author{Simon~Ekstr\"{o}m,
        Filip~Malmberg,
	H{\aa}kan~Ahlstr\"{o}m,
	Joel~Kullberg,
        and~Robin~Strand
\thanks{All authors are with the Department
of Surgical Sciences, Uppsala University, Sweden.}
\thanks{F. Malmberg and R. Strand are also with the Department of Information Technology, Uppsala University, Sweden.}
}
%
%

\markboth{Journal of \LaTeX\ Class Files,~Vol.~14, No.~8, August~2015}%
{Shell \MakeLowercase{\textit{et al.}}: Bare Demo of IEEEtran.cls for IEEE Journals}
%



\maketitle

\begin{abstract}
\emph{Objective:} Deformable image registration is a fundamental problem in medical image analysis, with applications such as longitudinal studies, population modeling, and atlas based image segmentation. Registration is often phrased as an optimization problem, i.e., finding a deformation field that is optimal according to a given objective function. Discrete, combinatorial, optimization techniques have successfully been employed to solve the resulting optimization problem. Specifically, optimization based on $\alpha$-expansion with minimal graph cuts has been proposed as a powerful tool for image registration. The high computational cost of the graph-cut based optimization approach, however, limits the utility of this approach for registration of large volume images. \emph{Methods:} Here, we propose to accelerate graph-cut based deformable registration by dividing the image into overlapping sub-regions and restricting the $\alpha$-expansion moves to a single sub-region at a time. \emph{Results:} We demonstrate empirically that this approach can achieve a large reduction in computation time -- from days to minutes -- with only a small penalty in terms of solution quality. \emph{Conclusion:} The reduction in computation time provided by the proposed method makes graph cut based deformable registration viable for large volume images. \emph{Significance:} Graph cut based image registration has previously been shown to produce excellent results, but the high computational cost has hindered the adoption of the method for registration of large medical volume images. Our proposed method lifts this restriction, requiring only a small fraction of the computational cost to produce results of comparable quality. 
\end{abstract}


%
\IEEEpeerreviewmaketitle

\section{Introduction}
Deformable image registration is a fundamental task in medical image analysis. Registration can be defined as the task of finding a spatial transformation that aligns two or more images with each other. We consider registration of two images -- one referred to as the \emph{source} image and the other referred to as the \emph{target} image. In this case, the deformable registration problem consists of finding a deformation field that indicates, for each voxel in the target image, the location of the corresponding voxel in the source image.  Applications of registration include fusion of multi-modal data, change detection and quantification, longitudinal studies, population modeling and studies of normal anatomical variability, and atlas based image segmentation~\cite{Survey,maintz1998}. 

Image registration methods can broadly be divided into two categories; \emph{parametric} and \emph{non-parametric}~\cite{Survey}. In parametric methods, the space of feasible transformations is reduced by introducing a parametrization (model) of the transformation. For example, the transformation may be required to be rigid or affine. Non-linear deformations may be parametrized using, e.g., B-splines, where a sparse grid of control points is used to specify a smooth deformation field. Here, we consider non-parametric, or \emph{dense} registration methods, where each point in the source image may be displaced arbitrarily, e.g. a separate displacement vector is stored for each voxel in the image. 

Dense registration is often phrased as an optimization problem, i.e., finding a deformation field that is optimal according to a given objective function which typically consist of two terms: a \emph{data} term measuring the degree to which the deformed source image resembles the target image, and a \emph{regularization} term enforcing the smoothness of the deformation field itself. 

By discretizing the set of allowed displacement vectors for a voxel, we may formulate dense deformable registration as a discrete labeling problem that can be solved using a variety of powerful inference methods. Here, we focus on the class of optimization methods referred to as \emph{move-making} algorithms. These algorithms start from an initial solution, and iteratively seek to find better solutions.  At each step, the algorithms search for a better solution in a limited subset of the combinatorial search space, referred to as the \emph{move-space}, and selects the one that leads to the largest reduction of the objective function.  The algorithms terminate when no such improving solution can be found. Provided that the move-space includes the current solution, these algorithms will converge in a finite number of steps. The resulting solution is then guaranteed to be locally optimal with respect to the move-space, in the sense that no better solution exists within the move-space of the current solution.

An important characteristic of a move-making algorithm is the size of its move-space. A large move space reduces the risk of getting stuck in poor local minima, and can lead to faster convergence of the algorithm. A prominent example of a move-making algorithm is the \emph{iterated conditional modes} (ICM) method, which iteratively attempts to improve the current solution by changing the label of a single element at a time, while keeping the others fixed. The move-space for ICM is thus small, and the algorithm is prone to getting stuck in poor local minima. Another move-making algorithm is the \emph{$\alpha$-expansion} method by Boykov et al.~\cite{boykov2001fast}. This algorithm selects a label $\alpha$ and considers moves that allow all elements to either keep their current label or change label to $\alpha$. Under certain conditions, the best such move can be determined in low-order polynomial time time by solving a max flow/min cut problem on a suitably constructed graph. Compared to ICM, the $\alpha$-expansion algorithm has a much larger move-space since it allows the labels of multiple elements to be changed simultaneously. In practice, this leads to much stronger local optima~\cite{szeliski2006comparative}. The drawback of the $\alpha$-expansion method is its high computational complexity.  Iteratively solving max flow/min cut problems on large graphs is very computationally expensive, and the computational cost grows non-linearly with respect to the image size. The high computational cost limits the practical utility of this approach for registration of large volume images~\cite{szmul2016graph}. Here, we propose an approach for reducing the computation time of graph cut based registration, while still producing high quality solutions.  

We may view the ICM and $\alpha$-expansion algorithms as two extremes in terms of their move-spaces; ICM only allows updates to a single variable at a time, while $\alpha$-expansion allows simultaneous updates to all variables.  While the ability to simultaneously update the labels of many voxels yields superior results, we hypothesize here that interactions between voxels that are very far away from each other are unlikely to affect the result significantly. Based on this hypothesis, we introduce a novel move-making strategy that explores the middle-ground between ICM and $\alpha$-expansion: we divide the image into small overlapping sub-regions, and define the move-space as the set of solutions that can be obtained by an $\alpha$-expansion within a single sub-region at a time. The optimal move within a sub-region can still be determined by solving a minimal graph cut problem, with special care taken to correctly handle the borders of each subregion. Selecting the size of the sub-regions involves a trade-off between computation time on one hand and solution quality on the other. We demonstrate empirically that this trade-off is benign, in the sense that a large reduction in computation time can be achieved with only a small penalty in terms of solution quality. The reduction in computation time makes graph cut based deformable registration viable for large volume images. Experiments are performed on registration of whole-body fat-water magnetic resonance (MR) images.




\section{Related work}
As shown by Boykov et al.~\cite{boykov2001fast}, minimal graph cuts are a powerful tool for solving dicrete optimization problems arising in image analysis and computer vision. The use of minimal graph cuts for deformable image registration was, to our knowledge, first proposed by Tang et al.~\cite{tang2007non}. The use of graph cuts in this context has been shown to produce excellent results, but the high computational cost associated with this method has limited its use in medical imaging. In recent years, the topic has since then been revisited by several authors. Heinrich~\cite{heinrich2013mrf}, formulate a similar optimization problem, but then prune the graph to a tree on which the optimal labeling (to an approximation of the original problem) can be found by efficient belief propagation. Szmul et al.~\cite{szmul2016graph, szmul2017supervoxels} proposed to accelerate graph-cut based image registration by partitioning the image into supervoxels to yield a sparse graph-based image representation. This reduces computation time, but the result is restricted by the assumption that the displacement field is constant within each supervoxel. Glocker et al.~\cite{glocker2008dense,glocker2011deformable} consider optimization problems similar to that presented here, but use a linear programming solver to find optimal solutions.


\section{Preliminaries}
\subsection{Notation}
We define a scalar volume image $\mathbf{I}$ as a pair $(V, I)$ consisting of a set $V$ of voxels (represented by points in $\mathbb{R}^3$) and a mapping $I : V \rightarrow \mathbb{R}$ where the value $I(v)$ of $I$ at $v \in V$ represents the image intensity at the voxel $v$. The voxels in $V$ are assumed to be arranged on a regular grid, so that each voxel can be identified by a coordinate in $\mathbb{Z}^3$. We let $\mathcal{N}$ denote the set of all adjacent voxels, i.e., $(v,w) \in \mathcal{N}$ if $v$ and $w$ are adjacent. Throughout, we will use the standard 6-neighborhood to define adjacency.

\subsection{Deformation model}

We denote the source image by $\mathbf{S}=(S,V_S)$ and the target image by $\mathbf{T}=(T,V_T)$. The two images are related by a transformation $W:\mathbb{R}^3 \rightarrow \mathbb{R}^3$. This transformation is assumed to map the target image to the source image, i.e., a so called backward transform~\cite{Survey}. With a backward transform this calculation is straightforward; for every voxel in the target image we simply interpolate the value at the corresponding location in the source image. The transformation at every voxel position $\mathbf{x}$ is given as the addition of an identity transform with a displacement field $\mathbf{u}$, i.e.,

\begin{equation}
\label{eq:disp}
W(\mathbf{x})=\mathbf{x}+\mathbf{u}(\mathbf{x}) \; .
\end{equation}

We store $\mathbf{u}$ for each voxel in the target image, and use tri-linear interpolation to determine the value of the deformation field at non-grid locations. Note that by Equation~\ref{eq:disp}, it is trivial to transform between $\mathbf{u}$ and $W$.

\section{Deformable registration by discrete optimization}
In this section, we formulate the deformable registration problem as a discrete optimization problem. 

\subsection{Matching criterion}
\label{sec:objective_function}
Matching criteria used for deformable image registration typically consist of two terms: a \emph{data} term measuring the degree to which the deformed source image resembles the target image and a \emph{regularization} term enforcing the smoothness of the deformation field itself. Many different data and regularization terms have been proposed in the literature, see, e.g.,~\cite{Survey} for an overview. In this section, we present the specific matching criterion used here for all experiments. We emphasize that our proposed efficient optimization scheme could also be used with other matching criteria, so long as they satisfy the criteria given in Section~\ref{sec:method}.

The images that are to be registered are assumed to be quantitative and of the same modality. Thus, the data term $D$ is simply taken to be the sum of squared differences (SSD) between the target image and the deformed source image. 

\begin{equation}
D(\mathbf{u})=\sum_{v\in V_T} |T(v)-S(v+\mathbf{u}(v))|^2 \; .
\end{equation}

\noindent For multi-channel images, the data term is taken to be the average SSD across all image channels.

For the regularization term $R$ we consider a \emph{diffusion} regularizer~\cite{glocker2008dense}, penalizing large first order derivatives in the deformation field: 

\begin{equation}
\label{eq:regularization}
R(\mathbf{u})=\sum_{(v,w)\in \mathcal{N}} \| \mathbf{u}(v)-\mathbf{u}(w)\|^2
\end{equation}

The total matching criterion $f$ is thus defined as 

\begin{equation}
\label{eq:energy_fn}
f(\mathbf{u})=(1-\alpha) D + \alpha R \; ,
\end{equation}

\noindent where $\alpha$ is a user defined parameter controlling the balance between the data and regularization terms. We seek a transformation $W'$ that minimizes $f(W)$.

\subsection{Optimization via minimal graph cuts}
\label{sec:GC_optmization}
By discretizing the space of possible deformation vectors, the problem of finding a deformation field minimizing the matching criterion $f$ becomes a discrete labeling problem. In this section, we review how such labeling problems can be solved via minimal graph cuts, employing variations on the $\alpha$-expansion method proposed by Boykov et al.~\cite{boykov2001fast}. The optimization procedure outlined below is computationally expensive, but forms the basis for the more efficient method proposed in Section~\ref{sec:method}.

Let $\mathbf{u}$ be a deformation field, and let $\delta \in \mathbb{R}^3$ be a vector. We say that a deformation field $\mathbf{u}'$ is \emph{$\delta$--adjacent} to $\mathbf{u}$ if $\mathbf{u}'$ can be obtained from $\mathbf{u}$ by adding the vector $\delta$ to the deformation vector at some of the voxels $v\in V_T$. Formally, $\mathbf{u}'$ is $\delta$--adjacent to $\mathbf{u}$ if it can be written as

\begin{equation}
\label{eq:delta_adjacent}
\mathbf{u}'(x)= \mathbf{u}(x) +L(x)\delta \; ,
\end{equation}

\noindent where $L$ is a binary labeling function that assigns to each voxel in $V_T$ either $0$ or $1$, i.e., $\mathbf{L}:V_T \rightarrow \{0,1\}$. Given a set $\Delta$ of 3D vectors, we say that $\mathbf{u}'$ is $\Delta$--adjacent to $\mathbf{u}$ if it is $\delta$--adjacent to $\mathbf{u}$ for some $\delta \in \Delta$. We say that a deformation field $\mathbf{u}$ is \emph{locally optimal} with respect to $\Delta$ if there is no $\Delta$--adjacent deformation field $\mathbf{u}'$ such that $f(\mathbf{u'})<f(\mathbf{u})$. In practice, we may relax this condition somewhat, and consider $\mathbf{u}$ to be locally optimal if there is no  $(\beta,\Delta)$--adjacent deformation field $\mathbf{u}'$ such that $f(\mathbf{u'})+\Upsilon<f(\mathbf{u})$ for some small, user defined tolerance value $\Upsilon$. A strategy for finding a deformation field that is locally optimal with respect to $\Delta$ can be defined as follows:

\begin{enumerate}
\item Start with an initial deformation field $\mathbf{u}$.
\item While there exists a deformation field $\mathbf{u}'$ that is  $\Delta$-adjacent to $\mathbf{u}$ such that $f(\mathbf{u}')<f(\mathbf{u}) + \Upsilon$, set $\mathbf{u} \leftarrow \mathbf{u}'$.
\end{enumerate} 

This procedure is an instance of a move-making optimization algorithm, where the move-space is given by all $\Delta$-adjacent solutions. Thus, the algorithm is guaranteed to terminate and return a locally optimal deformation field. The crucial issue in the above procedure is to determine, at every iteration of step 2, whether there exists a $\Delta$-adjacent deformation field to the current solution for which the matching criterion is improved. For any given deformation field, the set of $\Delta$-adjacent deformation fields are extremely large -- it has $|\Delta|2^{|V|}$ elements -- and so an exhaustive search of this set is not feasible. Instead, the $\Delta$-adjacent solution with the lowest matching criterion value is found by solving a minimal graph cut problem on a suitably constructed graph~\cite{kolmogorov2004energy}. 

For the specific matching criterion considered here we observe,  for a fixed $\mathbf{u}$ and $\delta\in \Delta$, that the value of the matching criterion for a $\delta$--adjacent deformation field $\mathbf{u}'$ can be expressed as a function of the labeling function $L$:

\begin{equation}
\label{eq:binary_labeling}
f(\mathbf{u}') = \sum_{v\in V_T} \phi_v(L(v)) + \sum_{(v,w) \in \mathcal{N}} \phi_{v,w}(L(v), L(w)) \; .
\end{equation}

In the above equation, the \emph{unary} terms $\phi_v$ represent the data and constraint term of the matching criterion, while the \emph{binary} terms $\phi_{v,w}$ represent the smoothness term. Formally, the unary terms are given by

\begin{equation}
\phi_v(0)=|T(v)-S(w+\mathbf{u}(v))|^2 \; , 
\end{equation}

\begin{equation}
\phi_v(1)=|T(v)-S(w+\mathbf{u}(v)+\delta)|^2  \; . 
\end{equation}

Similarly, the binary terms are given by

\begin{equation}
\label{eq:first_binary}
\phi_{v,w}(0,0)=\|\mathbf{u}(v)-\mathbf{u}(w)\|^2 \; ,
\end{equation}

\begin{equation}
\phi_{v,w}(1,1)=\|(\mathbf{u}(v) +\delta) -(\mathbf{u}(w) +\delta)\|^2 \; ,
\end{equation}

\begin{equation}
\phi_{v,w}(1,0)=\|(\mathbf{u}(v) +\delta) -\mathbf{u}(w) \|^2  \; , \mbox{and}
\end{equation}

\begin{equation}
\label{eq:last_binary}
\phi_{v,w}(0,1)=\|\mathbf{u}(v)-(\mathbf{u}(w)+\delta) \|^2  \; .
\end{equation}

We are interested in finding a labeling $L$ for which the matching criterion $f(\mathbf{u}')$ is minimal. As established by~\cite{kolmogorov2004energy} a globally optimal solution to binary labeling problems of the form given in Eq.~\ref{eq:binary_labeling} can be found by solving a maximum flow/minimum cut problem, provided that all binary terms are \emph{submodular}. In Appendix A, we give a proof that the binary terms given by Eq.~\ref{eq:first_binary}--\ref{eq:last_binary} are submodular for any $\mathbf{u}$ and $\delta$.

This result enables an efficient approach for determining if a given deformation field $\mathbf{u}$ is locally optimal with respect to $\Delta$: For every $\delta\in \Delta$, we solve a max-flow problem to find, among all deformation fields that are $\delta$-adjacent to $\mathbf{u}$, one for which $f$ is minimal. This results in $|\Delta|$ deformation fields. If the value of the matching criterion for any of these fields is smaller than $f(\mathbf{u})$, we accept the deformation field as the new current solution. If not, then $\mathbf{u}$ is locally optimal with respect to $\Delta$. 

To implement this approach in practice, we need to select a set of vectors $\Delta$. The number of vectors in $\Delta$ and their capture range has a significant effect on the registration process. Generally, a larger set of vectors allows the algorithm to ``explore'' a larger region around the current solution, potentially leading to stronger optima. On the other hand, we wish to keep the set of vectors small for efficient optimization. Here, we have opted for a small set of displacement vectors $\Delta$, representing small local displacements of the current deformation field along the main axis directions. Formally, we take $\Delta$ to be the set of vectors $\delta$ of the form

\begin{equation}
\delta = \pm \epsilon \mathbf{e}_i \; ,
\end{equation}

\noindent where $\epsilon \in \mathbb{R}$ is a user specified step length and $\mathbf{e}_i$ is one of the standard basis vectors $(1,0,0)$, $(0,1,0)$ and $(0,0,1)$. Typically, we set $\epsilon$ to a fraction of the voxel spacing, to allow for sub-voxel precision in the registration.

To find a good solution with such a small set $\Delta$, the algorithm must be initialized with a deformation field that is close to the desired one. To alleviate this requirement and increase the "capture range" of the registration, we follow the common practice of combining the move-making optimization with a multi-resolution strategy. Starting with the original image, we obtain a lower resolution image by applying a Gaussian filter at the current resolution, and then discarding every second voxel of the filtered image along each dimension. This procedure is repeated $k-1$ times to obtain $k$ resolution levels, where $k$ is specified by the user. Registration then starts at the coarsest resolution, and applies the optimization method described above to obtain a locally optimal transformation at the current resolution. The resulting deformation field is up-sampled using tri-linear interpolation to obtain a starting guess for the next finer resolution. The process is repeated until a locally optimal solution has been calculated at the finest resolution. 

\section{Proposed efficient optimization method}
\label{sec:method}
The cost of solving a single max flow/min cut problem grows non-linearly with the number of nodes in the graph. Thus, iteratively finding optimal adjacent solutions for the entire volume is computationally expensive. To reduce computational cost, we propose to divide the volume into smaller sub-regions, and restrict the move-space so that only voxels within the same sub-region are allowed to simultaneously update their displacement vectors within a single move. This is motivated by the hypothesis that interactions between voxels that are very far away from each other are unlikely to affect the result significantly.

Let $V'\subset V_T$ be a subset of the voxels in the target image. Given a deformation field $\mathbf{u}$ and a vector $\delta$, we say that a deformation field $\mathbf{u}'$ is $(V',\delta)$-adjacent to $\mathbf{u}$ if  $\mathbf{u}'$ can be obtained from $\mathbf{u}$ by adding the vector $\delta$ to the deformation vector at some of the voxels $v\in V'$. Formally, $\mathbf{u}'$ is  $(V',\delta)$-adjacent to $\mathbf{u}$ if it can be written on the form given in Eq.~\ref{eq:delta_adjacent}, with the additional condition that $L(v)=0$ for all voxels $v\notin V'$.

Given a set $\Delta$ of 3D vectors and set $\beta$ of subsets of $V_T$, we say that $\mathbf{u}'$ is $(\beta,\Delta)$--adjacent to $\mathbf{u}$ if it is $(V',\delta)$-adjacent to $\mathbf{u}$ for some $\delta \in \Delta$ and $V'\in \beta$. We say that a deformation field $\mathbf{u}$ is \emph{locally optimal} with respect to $(\beta,\Delta)$ if there is no $(\beta,\Delta)$--adjacent deformation field $\mathbf{u}'$ such that $f(\mathbf{u'})<f(\mathbf{u}) + \Upsilon$. 

The move-making strategy for finding a deformation field that is locally optimal with respect to $(\beta,\Delta)$ can be defined as follows:

\begin{enumerate}
\item Start with an initial deformation field $\mathbf{u}$.
\item While there exists a deformation field $\mathbf{u}'$ that is $(\beta,\Delta)$-adjacent to $\mathbf{u}$ such that $f(\mathbf{u}')+\Upsilon<f(\mathbf{u})$, set $\mathbf{u} \leftarrow \mathbf{u}'$.
\end{enumerate} 

As before, we use minimal graph cuts to efficiently search the set of $(\beta,\Delta)$--adjacent deformation fields for a solution with better matching criterion value. Let $\mathcal{N}'$ be the set of adjacent voxel pairs in $\mathcal{N}$ for which at least one of the voxels in the pair is in $V'$. For a fixed $\mathbf{u}$, voxel subset $V'\in \beta$ and vector $\delta$, the value of the matching criterion for a $(V',\delta)$-adjacent deformation field $\mathbf{u}'$ can again be expressed as a function of the labeling function $L$:

\begin{equation}
\label{eq:binary_sublabeling}
\begin{split}
E(L)= &\sum_{v\in V_T \setminus{V'}} \phi_v(0) +\sum_{(v,w)\in \mathcal{N}\setminus{\mathcal{N}'} } \phi_{v,w}(0,0) +  \\ 
&\sum_{v\in V'} \phi_v(L(v)) +  \sum_{(v,w)\in \mathcal{N}' } \phi_{v,w}(L(v),L(w)) \; .
\end{split}
\end{equation}

The first two terms in the above equation are constant with respect to $L$ and thus do not affect the optimization of $L$. Let $\mathcal{N}''$ be the set of adjacent voxel pairs in $\mathcal{N}$ for which \emph{both} voxels in the pair are in $V'$. We can then rewrite the remaining terms in Eq.~\ref{eq:binary_sublabeling} as 

\begin{equation}
\label{eq:subproblem}
\begin{split}
\sum_{v\in V'} \phi_v(L(v)) +  \sum_{(v,w)\in \mathcal{N}'' } \phi_{v,w}(L(v),L(w))+ \\ \sum_{(v,w)\in \mathcal{N}'\setminus \mathcal{N}'' } \phi_{v,w}(L(v),0)
\end{split}
\end{equation}

\noindent In the last term of the above equation, we make the assumption that $w$ is the voxel in the pair $(v,w)$ that is not in $V'$. This last term is a sum over binary terms, but since one of the labels, $L(w)$, is fixed to $0$ we can treat each $\phi_{v,w}(L(v),0)$ as a unary term for the remaining label $L(v)$. Thus a labeling $L$ that globally minimizes Eq.~\ref{eq:subproblem} can be found by computing a minimal graph cut. Pseudo-code for the full proposed optimization algorithm is listed in Algorithm~\ref{alg:optimization}. 

\begin{algorithm}
\caption{Optimization algorithm}
\label{alg:optimization}
\begin{algorithmic}
\STATE{\textbf{Input:} An initial deformation field $\mathbf{u}$. A set of vectors $\Delta$, a set of voxel subsets $\beta$, and a tolerance value $\Upsilon$.}
\STATE{\textbf{Output:} A locally optimal deformation field $\mathbf{u}$}
\STATE{$\mathbf{u}(v)=0$ for all $v\in V$\;}
\STATE{done $=$ false\;}
\WHILE{done $=$ false}
\STATE{done $=$ true\;}
\FORALL{$V' \in \beta$}
\FORALL{$\delta \in \Delta$}
  \STATE{
  Solve a min-cut/max-flow problem to find, among all deformation fields that are $(V', \delta)$-adjacent to $\mathbf{u}$, a deformation field $\mathbf{u}'$ for which $f(\mathbf{u}')$ is minimal.\;
  \IF{$f(\mathbf{u}') + \Upsilon< f(\mathbf{u})$}
  \STATE{
    Set $\mathbf{u} $=$ \mathbf{u}'$ \; 
    Set done $=$ false\;
    }
  \ENDIF
  }
\ENDFOR
\ENDFOR
\ENDWHILE
\end{algorithmic}
\end{algorithm}

To select the set of voxel subsets $\beta$ in practice, we consider a grid of rectangular blocks covering the target image. Each block $B$, has a size of $n \times n \times n$ voxels. Taking $\beta=B$ can potentially lead to artifacts at the boundaries between the blocks. To avoid this, we also consider a second set of blocks, $B_\mathrm{shift}$, with blocks of the same size as in $B$ but translated by $n/2$ along each dimension, and take $\beta=B \cup B_\mathrm{shift}$. The size $n$ of the blocks offers a trade-off between computational cost and solution quality. If $n$ is too small, the optimization algorithm will tend to get stuck in poor local optima. Selecting a too large $n$, on the other hand, may result in very long computation times but not much gain in solution quality. In Section~\ref{sec:experiments}, we explore this trade-off empirically. 

\subsection{Computational optimizations}
The move-making optimization procedure described above divides the voxels of the target volume into a set of overlapping rectangular blocks $\beta$. The algorithm then repeatedly iterates over all blocks, and tries to improve the current solution by modifying the displacement vectors within a single block at a time by solving a graph cut problem. Reducing the size of the graph cut problems to be solved during optimization greatly reduces the computation time. The division into subregions also facilitates some additional computational optimizations, described in this section, that further accelerate the algorithm.

To solve each graph cut problem, we use the fast graph cut algorithm proposed by Boykov et al.~\cite{boykov2004experimental}. As can be observed in Eqs.~\ref{eq:binary_sublabeling} and~\ref{eq:subproblem}, the optimization of a single block $V'$ does not depend on the current deformation vectors at all target image voxels. In fact, the optimization of a given block $V' \in \beta$ only depends on the deformation vector at a voxel $v$ if

\begin{enumerate}
\item $v$ is in $V'$, or
\item $v$ is adjacent to a voxel in $V'$.
\end{enumerate}

\noindent We denote by $\Gamma (V')$ the set of voxels in $V_T$ satisfying one of the above criteria. If, for two blocks $V'$ and $V''$, it holds that $\Gamma(V') \cap \Gamma(V'') \ne \emptyset$, then we say that $V'$ and $V''$ are \emph{dependent}.

\begin{figure}[t]
\centering
\includegraphics[width=0.5\columnwidth]{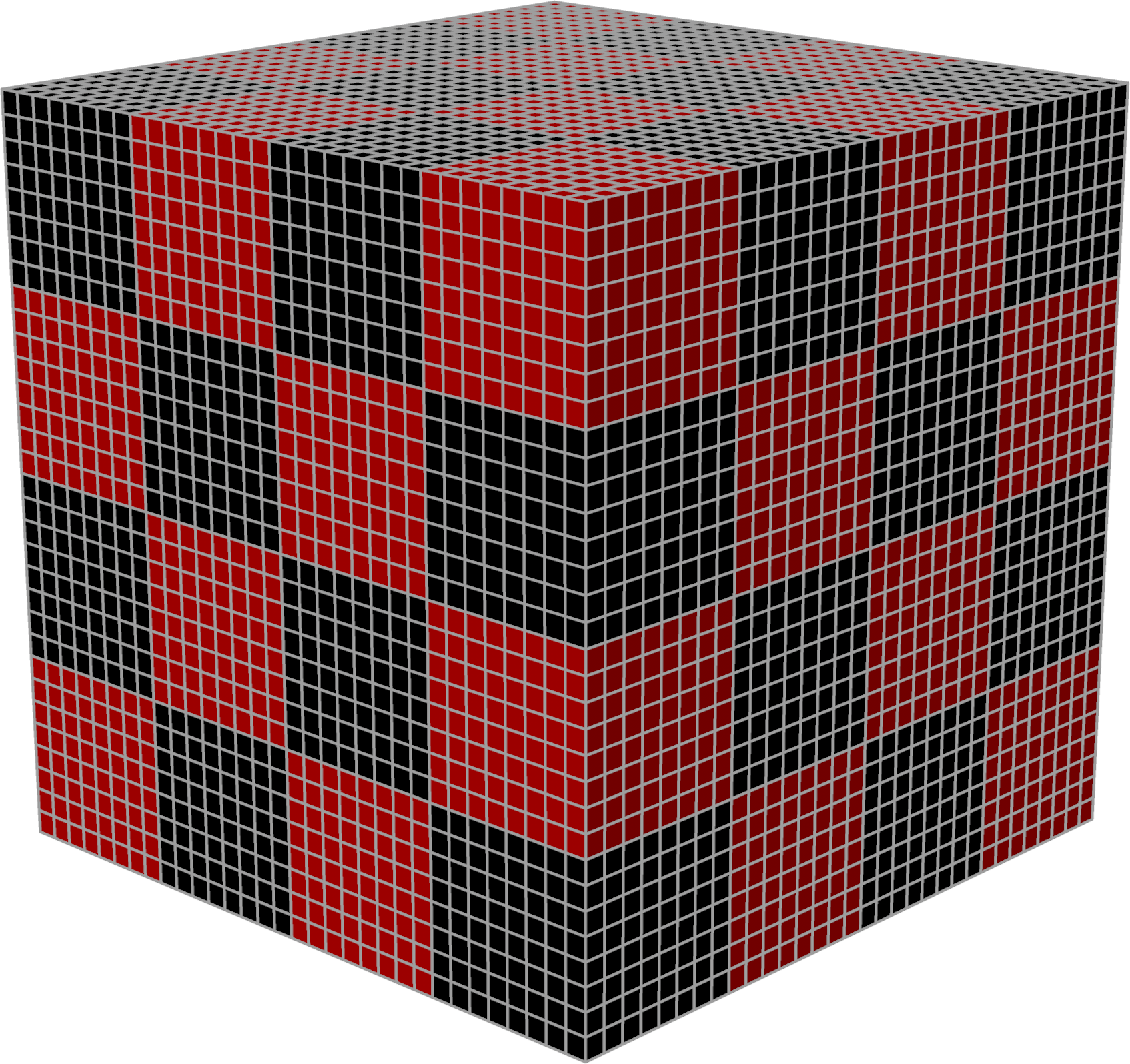}
\caption{Illustration of red-black ordering of blocks. In this illustration, each of the $4 \times 4 \times 4$ blocks contains $n \times n \times n =8 \times 8 \times 8$ voxels.}
\label{fig:red_black}
\end{figure}

The first computational optimization is to parallelize the iteration over the voxel subsets in $\beta$. Recall that we define $\beta$ as the union of two grids of rectangular blocks $B$ and $B_\mathrm{shift}$, translated relative to each other by half the block size in each dimension.  We arrange the loop over $\beta$ in Algorithm~\ref{alg:optimization} so that we first iterate over all blocks in $B$, and then over all blocks in $B_\mathrm{shift}$. We observe that when iterating over a single set of blocks $B' \in \{B,B_\mathrm{shift}\}$, two blocks $V'$ and $V''$ in $B'$ are dependent precisely if they are adjacent, i.e., there is a pair of adjacent voxels in $\mathcal{N}$ for which one voxel is in $V'$ and the other is in $V''$. This allows us to use a \emph{red-black ordering} to parallelize the iteration over a single set of blocks; the set of blocks $B'$ is further divided into two groups, identified as \emph{red} and \emph{black} blocks, such that black blocks are only adjacent to -- and thus dependent on --red blocks, and vice versa. See Fig.~\ref{fig:red_black}. All black blocks can then be updated independently, as can all red blocks. Our implementation uses OpenMP\footnote{http://openmp.org} to distribute the work among multiple CPU cores.

The second computational optimization is based on the observation that after a few iterations over the blocks, most blocks have already converged to their locally optimal state, and remain unchanged for subsequent iterations. Thus, a large amount of redundant work is spent on trying to improve the deformation field within blocks that have already reached a locally optimal state. Specifically, if a block $V'$ was not changed at one iteration, it is guaranteed to remain unchanged also at the next iteration unless a dependent block $V'' \in \beta$ was changed since the last update of $V'$. Keeping track of these conditions allows to implement \emph{early termination}, i.e., not performing the graph cut optimization step for blocks that will not be changed anyway.

\section{Experiments and results}
\label{sec:experiments}
Several previous studies have established the usefulness of graph-cut based optimization for deformable image registration~\cite{tang2007non,so2011non}. Thus, the focus of this section is not to study the performance of graph-cut based registration per se. Instead we focus on evaluating how the proposed division into subregions affects the optimality of the result.  We evaluate the performance of the proposed optimization method on a set of eleven whole body MR images of female subjects from the \emph{POEM} cohort (\url{www.medsci.uu.se/poem}). 

All subjects were imaged on a 1.5T clinical MRI system (Philips Achieva, Philips Healthcare, Best, Netherlands) in supine position using the body coil and a whole body water-fat imaging protocol that used a spoiled 3D multi gradient echo sequence. Scan parameters were: TR/TE1/$\Delta$TE = 5.9/1.36/1.87 ms, 3 unipolar echoes, flip angle 3. Imaged field of view (FOV) 530 mm $\times$ 377 mm $\times$ 2000 mm, reconstructed voxel size 2.07 mm $\times$ 2.07 mm $\times$ 8.0 mm in sagittal$\times$coronal$\times$axial directions. The imaging protocol and the water-fat image reconstruction have been described previously~\cite{berglund2010three,kullberg2009automated}.

Fat- and water fraction volume images were calculated from the MR image data as described by~\cite{berglund2010three}. Background noise outside the body was suppressed by multiplication with a body mask, obtained using the approach described by~\cite{strand2017concept}. We treat each pair of fat- and water fraction images as a single two-channel volume image, and use these as input to the registration algorithm. 

One subject was selected as the reference subject, to which the other ten subjects were registered. The registration parameters used were a step length of $\epsilon = 0.5$ (specified in image coordinates), $k=6$ resolution levels in the multiresolution pyarmid, regularization parameter $\alpha=0.1$, and the tolerance level $\Upsilon$ was set to $10^{-5}$. All pairs were registered in both forward and reverse direction. The quality of the results was assessed using the value of the matching criterion $f(\mathbf{u})$ of the final deformation field and the inverse consistency \emph{vector magnitude error} (VME)\cite{christensen2001consistent}, computed as 

\begin{equation}
  \label{eq:ic_vme}
  VME=\frac{1}{|V|}\sum_{x\in V} |x - T_{B \rightarrow A} \circ T_{A \rightarrow B}(x)|,
\end{equation}
where $T_{B \rightarrow A}(T_{A \rightarrow B}(x))$ is the composite of the deformation fields in forward and reverse directions.

\begin{figure*}[tbh]
  \centering
  \includegraphics[width=\textwidth]{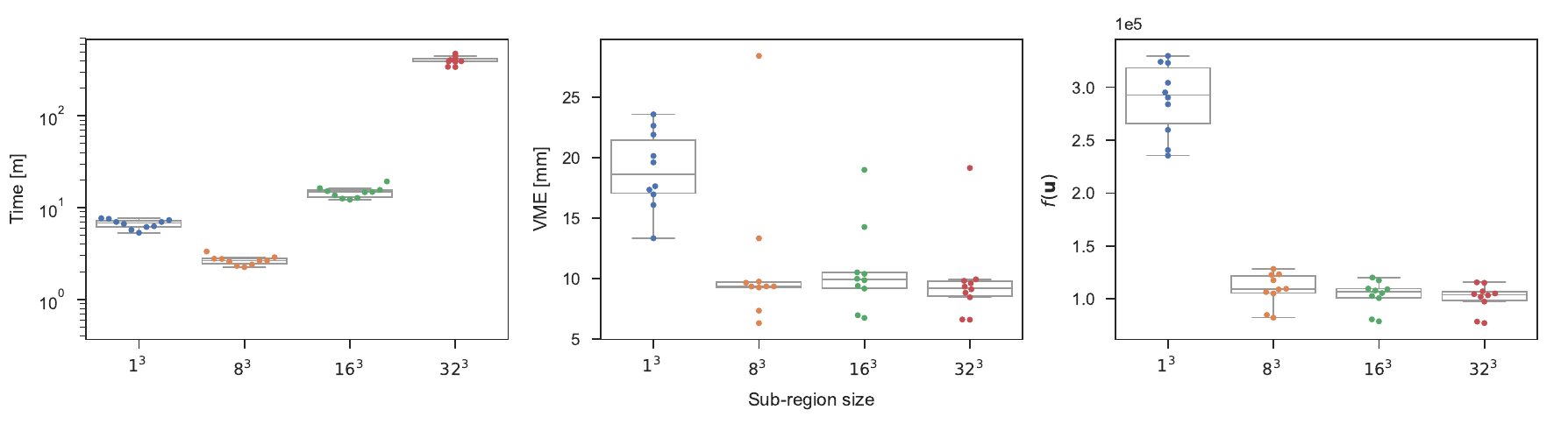}
  \caption{Total computation time, VME, and resulting matching criterion $f(\mathbf{u})$ presented for registrations using sub-region size parameters $1^3, 8^3$, $16^3$, and $32^3$. Note that the Y-axis in the time plot is shown in logarithmic scale. }
  \label{fig:level0_time_vme_energy}
\end{figure*}

 Registrations were performed on the 10 subject pairs with varying size on the sub-regions and compared to registration without sub-regions, i.e. direct $\alpha$-expansion. Block shaped sub-regions of size $\kappa \in \{1^3, 8^3, 16^3, 32^3\}$ were used. During the experiment, the computation time for registration with direct $\alpha$-expansion was shown to exceed 5 days per subject and for practical reasons these computations were thus terminated. The results of the remaining experiments are shown in Figure~\ref{fig:level0_time_vme_energy}. As the figure shows, the computation time increases dramatically as we increase in the sub-region size. In terms of VME and matching criterion, $f(\mathbf{u})$, the effect is less noticeable. Both measures are drastically improved as we move from a sub-region size of $1^3$ (equivalent to ICM) to a size of $8^3$, but increasing the sub-region size further only affects the VME and matching criterion marginally. 

\setlength\tabcolsep{1.5pt} 
\begin{figure}
  \centering
  \begin{tabular}{ccccc}
    \begin{subfigure}{0.185\columnwidth}
      \includegraphics[width=\textwidth]{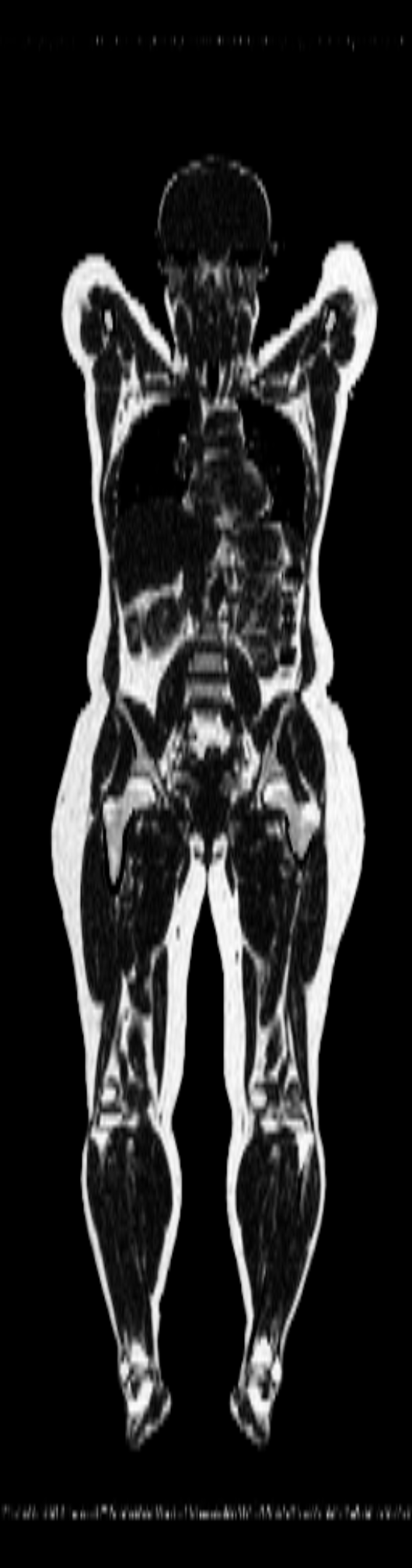}
      \caption{}
    \end{subfigure} &
    \begin{subfigure}{0.185\columnwidth}
      \includegraphics[width=\textwidth]{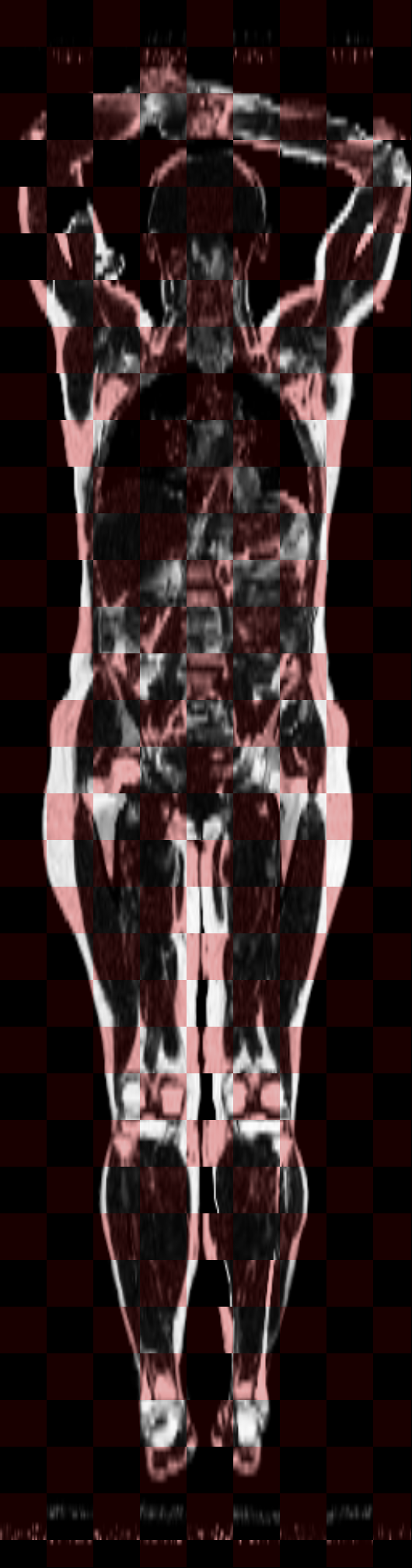}
      \caption{}
    \end{subfigure} & 
    \begin{subfigure}{0.185\columnwidth}
      \includegraphics[width=\textwidth]{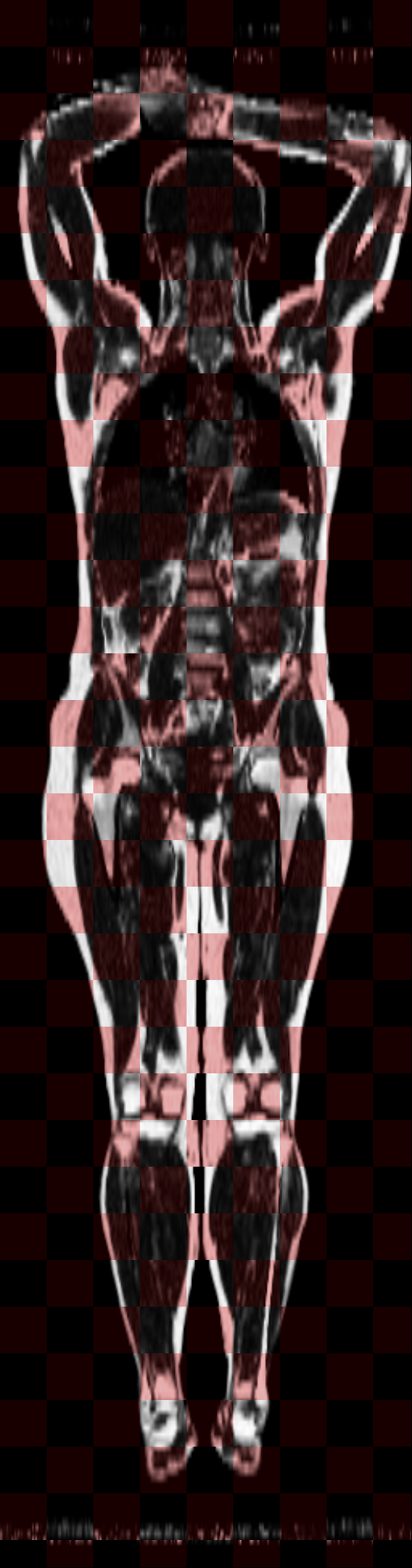}
      \caption{}
    \end{subfigure} &
    \begin{subfigure}{0.185\columnwidth}
      \includegraphics[width=\textwidth]{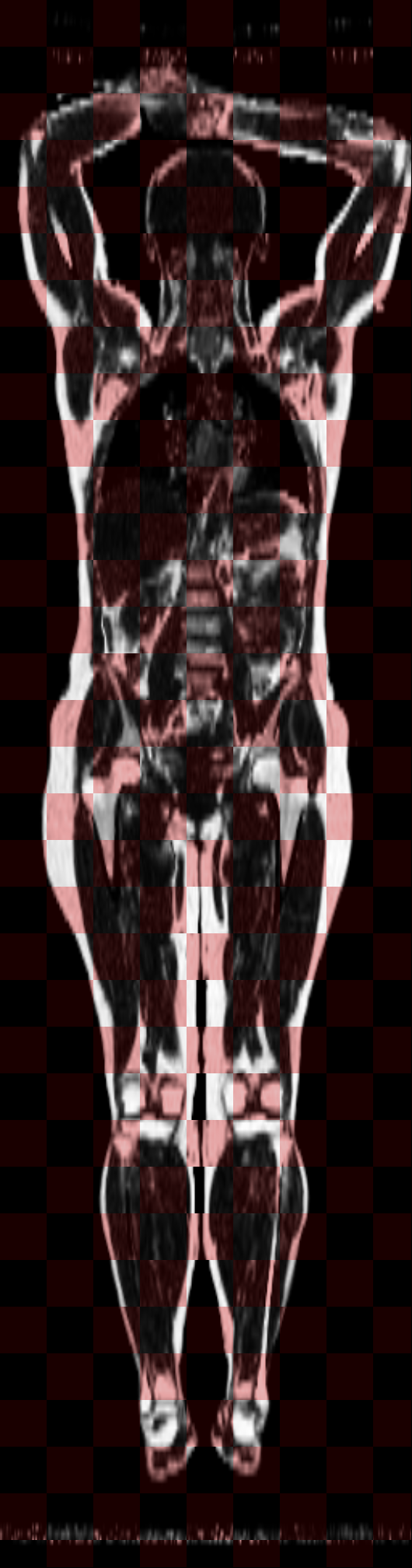}
      \caption{} 
    \end{subfigure} &
    \begin{subfigure}{0.185\columnwidth}
      \includegraphics[width=\textwidth]{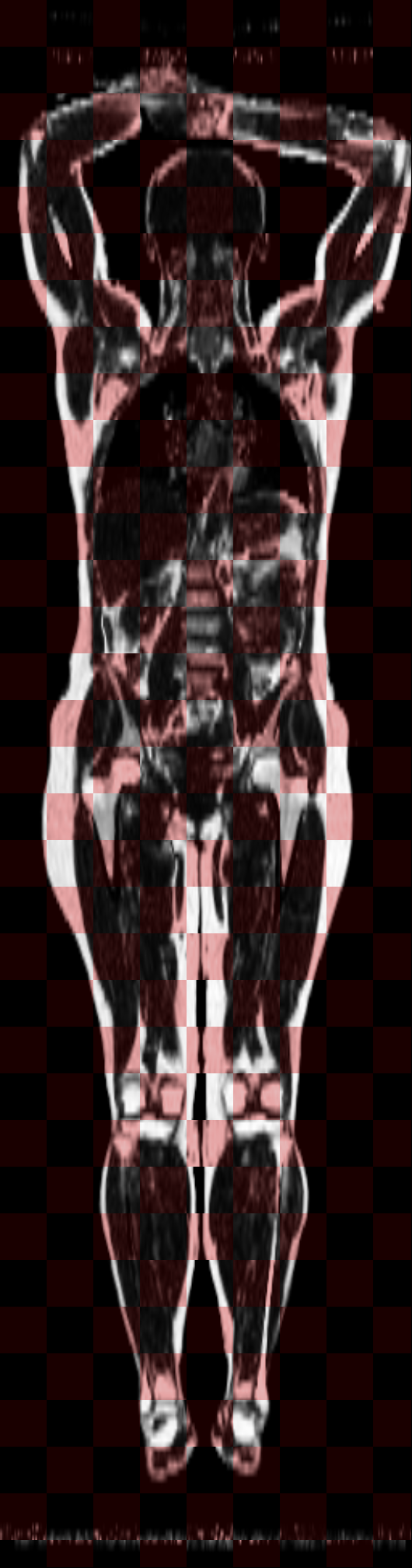}
      \caption{}
    \end{subfigure}
  \end{tabular}
  \caption{Coronal slices from the fat percent images from one of the subjects registered in the experiment. Presented are the unregistered fat percent image (a) and checkerboards showing the registered fat images (white) and the reference images (red) for sub-region sizes $1^3$ (b), $8^3$ (c), $16^3$ (d), and $32^3$ (e).}
  \label{fig:reg_result}
\end{figure}

Fig.~\ref{fig:reg_result} shows a sample from the registered subjects from the experiment. Presented are the original unregistered fat percent image and checkboards showing the registered fat images (white) and the references images (red) for the various sub-region sizes. The checkerboards visualizes the similarity between the registered subject and the reference subject. Visually, these results agree with the VME and matching criterion results. The registration result with sub-region size $1^3$ (b) exhibits visual errors throughout the image, most notably in arms and lower legs. For the other sub-region sizes (c,d,e) there were small errors around the calves but none of the other large errors from (b) are visible. In comparison, there are no evident differences between (c,d,e), indicating that the choice of sub-region size, within a reasonable range, have little or no effect on the visual end result.

\begin{figure*}[tbh]
  \centering
  \includegraphics[width=\textwidth]{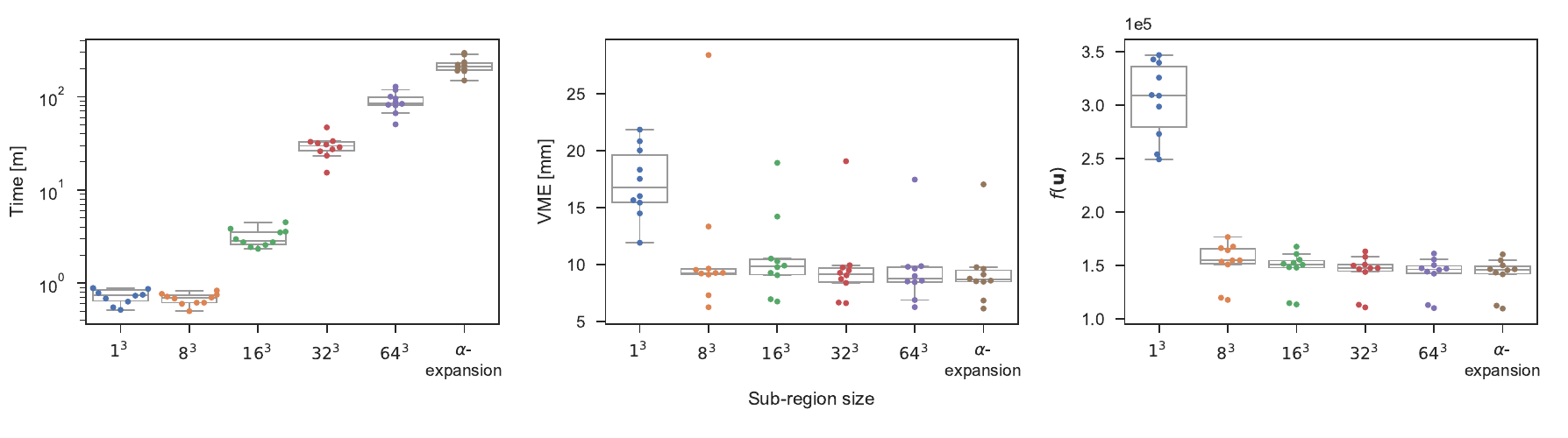}
  \vspace*{-7.5mm}
  \caption{Total computation time, VME, and resulting matching criterion $f(\mathbf{u})$ for registrations using sub-region size parameters $1^3$, $8^3$, $16^3$, $32^3$, $64^3$, and registrations with direct $\alpha$-expansion, with image resolution halved along each dimension. Note that the Y-axis in the time plot is shown in logarithmic scale.  }
  \label{fig:time_vme_energy}
\end{figure*}

To facilitate a comparison between our proposed method and direct $\alpha$-expansion, we repeated the experiment but performed the registration to the second-last level in the resolution pyramid, i.e., we register the volumes downsampled by a factor two along each dimension. In this experiment, registrations were performed using sub-regions of size $\kappa \in \{1^3, 8^3, 16^3, 32^3, 64^3\}$, as well as using direct $\alpha$-expansion. The results of this second experiment are shown in Figure~\ref{fig:time_vme_energy}. The results are in agreement with the previous experiment; computation time increases dramatically with increased sub-region size, but the matching criterion and VME are not substantially decreased for sub-region sizes beyond $8^3$. Paired two-sided t-tests comparing the results for sub-region size $8^3$ and direct $\alpha$-expansion resulted in $p=0.4$ for VME and $p=0.24$ for the matching criterion, telling us there's no significant difference between the two configurations. 

All registrations were performed on an Intel i7 2600K (3.4 GHz) with 4 cores and hyperthreading enabled (8 hardware threads).

\section{Conclusion} 
Graph-cut based optimization with $\alpha$-expansion has previously been shown to be a powerful tool for deformable image registration. The high computational cost of this approach, however, has limited the utility of this approach for registration of large volume images. The high computational cost was illustrated in our experiments, where registration time of a single pair of images by direct application of $\alpha$-expansion exceeded five days. We have presented an approach for fast discrete optimization for deformable registration of volume images. By dividing the image into subregions and restricting the $\alpha$-expansion moves to a single sub-region at a time, we achieve a large reduction in computation time --from days to minutes -- with only a small penalty in terms of solution optimality. The division into subregions also facilitates additional computational optimizations, parallelization and early termination, that both reduce the computation time substantially. 

We have demonstrated the utility of the proposed method in a case study on registration of whole-body fat-water MRI images, and conclude that the reduction in computation time facilitated by the proposed method makes graph-cut based optimization a viable option for registration of large volume images.

\appendix[Proof of submodularity]
A globally optimal solution to the binary labeling problem given in Eq.~\ref{eq:binary_labeling} can be found by solving a maximum flow/minimum cut problem, provided that all binary terms are \emph{submodular}~\cite{kolmogorov2004energy}. In this appendix, we prove that the binary terms in Eq.~\ref{eq:binary_labeling}, given by Eq.~\ref{eq:first_binary}--\ref{eq:last_binary}, are submodular for any $\mathbf{u}$ and $\delta$. 

A binary term $\phi_{v,w}(L(v),L(w))$ is \emph{submodular} if it satisfies the inequality

\begin{equation}
\label{eq:submodularity}
\phi_{v,w}(0,0)+\phi_{v,w}(1,1) \leq \phi_{v,w}(0,1)+\phi_{v,w}(1,0) \; .
\end{equation}

\noindent Here, we have 

\begin{equation}
\phi_{v,w}(0,0)=\|\mathbf{u}(v)-\mathbf{u}(w)\|^2 \; ,
\end{equation}

\begin{equation}
\phi_{v,w}(1,1)=\|(\mathbf{u}(v) +\delta) -(\mathbf{u}(w) +\delta)\|^2 \; ,
\end{equation}

\begin{equation}
\label{eq:10}
\phi_{v,w}(1,0)=\|(\mathbf{u}(v) +\delta) -\mathbf{u}(w) \|^2  \; ,
\end{equation}

\begin{equation}
\label{eq:01}
\phi_{v,w}(0,1)=\|\mathbf{u}(v)-(\mathbf{u}(w)+\delta) \|^2  \; .
\end{equation}

Noting that $\phi_{v,w}(0,0)= \phi_{v,w}(1,1)$, the left hand side of Eq.~\ref{eq:submodularity} can be rewritten as

\begin{equation}
\phi_{v,w}(0,0)+\phi_{v,w}(1,1) = 2\|(\mathbf{u}(v) -\mathbf{u}(w)\|^2 \; 
\end{equation}

The right hand side of Eq.~\ref{eq:submodularity} can be rewritten as

\begin{equation}
\phi_{v,w}(0,1)+\phi_{v,w}(1,0) = 2\|\mathbf{u}(v) -\mathbf{u}(w)\|^2+2\|\delta\|^2
\end{equation}

\noindent Since $2\|\delta\|^2\geq 0$, we have

\begin{equation}
\phi_{v,w}(0,0)+\phi_{v,w}(1,1) \leq \phi_{v,w}(0,1)+\phi{v,w}(1,0) \; .
\end{equation}

\noindent This completes the proof.


%



\section*{Acknowledgment}
Funding was received from Swedish Research Council (2016-01040).

\ifCLASSOPTIONcaptionsoff
  \newpage
\fi



\bibliographystyle{IEEEtran}
\bibliography{references}

\begin{thebibliography}{10}
\providecommand{\url}[1]{#1}
\csname url@samestyle\endcsname
\providecommand{\newblock}{\relax}
\providecommand{\bibinfo}[2]{#2}
\providecommand{\BIBentrySTDinterwordspacing}{\spaceskip=0pt\relax}
\providecommand{\BIBentryALTinterwordstretchfactor}{4}
\providecommand{\BIBentryALTinterwordspacing}{\spaceskip=\fontdimen2\font plus
\BIBentryALTinterwordstretchfactor\fontdimen3\font minus
  \fontdimen4\font\relax}
\providecommand{\BIBforeignlanguage}[2]{{%
\expandafter\ifx\csname l@#1\endcsname\relax
\typeout{** WARNING: IEEEtran.bst: No hyphenation pattern has been}%
\typeout{** loaded for the language `#1'. Using the pattern for}%
\typeout{** the default language instead.}%
\else
\language=\csname l@#1\endcsname
\fi
#2}}
\providecommand{\BIBdecl}{\relax}
\BIBdecl

\bibitem{Survey}
A.~Sotiras, C.~Davatzikos, and N.~Paragios, ``Deformable medical image
  registration: A survey,'' \emph{IEEE Transactions on Medical Imaging},
  vol.~32, no.~7, pp. 1153--1190, 2013.

\bibitem{maintz1998}
J.~A. Maintz and M.~A. Viergever, ``A survey of medical image registration,''
  \emph{Medical Image Analysis}, vol.~2, no.~1, pp. 1--36, 1998.

\bibitem{boykov2001fast}
Y.~Boykov, O.~Veksler, and R.~Zabih, ``Fast approximate energy minimization via
  graph cuts,'' \emph{IEEE Transactions on Pattern Analysis and Machine
  Intelligence}, vol.~23, no.~11, pp. 1222--1239, 2001.

\bibitem{szeliski2006comparative}
R.~Szeliski, R.~Zabih, D.~Scharstein, O.~Veksler, V.~Kolmogorov, A.~Agarwala,
  M.~Tappen, and C.~Rother, ``A comparative study of energy minimization
  methods for markov random fields,'' in \emph{European Conference on Computer
  Vision}.\hskip 1em plus 0.5em minus 0.4em\relax Springer, 2006, pp. 16--29.

\bibitem{szmul2016graph}
A.~Szmul, B.~W. Papiez, R.~Bates, A.~Hallack, J.~A. Schnabel, and V.~Grau,
  ``Graph cuts-based registration revisited: A novel approach for lung image
  registration using supervoxels and image-guided filtering,'' in
  \emph{Proceedings of the IEEE Conference on Computer Vision and Pattern
  Recognition Workshops}, 2016, pp. 152--159.

\bibitem{tang2007non}
T.~W. Tang and A.~C. Chung, ``Non-rigid image registration using graph-cuts,''
  in \emph{Medical Image Computing and Computer-Assisted Intervention--MICCAI
  2007}.\hskip 1em plus 0.5em minus 0.4em\relax Springer, 2007, pp. 916--924.

\bibitem{heinrich2013mrf}
M.~P. Heinrich, M.~Jenkinson, M.~Brady, and J.~A. Schnabel, ``{MRF}-based
  deformable registration and ventilation estimation of lung {CT},'' \emph{IEEE
  Transactions on Medical Imaging}, vol.~32, no.~7, pp. 1239--1248, 2013.

\bibitem{szmul2017supervoxels}
A.~Szmul, B.~W. Papie{\.z}, A.~Hallack, V.~Grau, and J.~A. Schnabel,
  ``Supervoxels for graph cuts-based deformable image registration using guided
  image filtering,'' \emph{Journal of Electronic Imaging}, vol.~26, no.~6, p.
  061607, 2017.

\bibitem{glocker2008dense}
B.~Glocker, N.~Komodakis, G.~Tziritas, N.~Navab, and N.~Paragios, ``Dense image
  registration through {MRF}s and efficient linear programming,'' \emph{Medical
  Image Analysis}, vol.~12, no.~6, pp. 731--741, 2008.

\bibitem{glocker2011deformable}
B.~Glocker, A.~Sotiras, N.~Komodakis, and N.~Paragios, ``Deformable medical
  image registration: Setting the state of the art with discrete methods,''
  \emph{Annual Review of Biomedical Engineering}, vol.~13, pp. 219--244, 2011.

\bibitem{kolmogorov2004energy}
V.~Kolmogorov and R.~Zabin, ``What energy functions can be minimized via graph
  cuts?'' \emph{IEEE Transactions on Pattern Analysis and Machine
  Intelligence}, vol.~26, no.~2, pp. 147--159, 2004.

\bibitem{boykov2004experimental}
Y.~Boykov and V.~Kolmogorov, ``An experimental comparison of min-cut/max-flow
  algorithms for energy minimization in vision,'' \emph{IEEE Transactions on
  Pattern Analysis and Machine Intelligence}, vol.~26, no.~9, pp. 1124--1137,
  2004.

\bibitem{so2011non}
R.~W. So, T.~W. Tang, and A.~C. Chung, ``Non-rigid image registration of brain
  magnetic resonance images using graph-cuts,'' \emph{Pattern Recognition},
  vol.~44, no.~10, pp. 2450--2467, 2011.

\bibitem{berglund2010three}
J.~Berglund, L.~Johansson, H.~Ahlstr{\"o}m, and J.~Kullberg, ``Three-point
  {Dixon} method enables whole-body water and fat imaging of obese subjects,''
  \emph{Magnetic Resonance in Medicine}, vol.~63, no.~6, pp. 1659--1668, 2010.

\bibitem{kullberg2009automated}
J.~Kullberg, L.~Johansson, H.~Ahlstr{\"o}m, F.~Courivaud, P.~Koken, H.~Eggers,
  and P.~B{\"o}rnert, ``Automated assessment of whole-body adipose tissue
  depots from continuously moving bed {MRI}: A feasibility study,''
  \emph{Journal of Magnetic Resonance Imaging}, vol.~30, no.~1, pp. 185--193,
  2009.

\bibitem{strand2017concept}
R.~Strand, F.~Malmberg, L.~Johansson, L.~Lind, M.~Sundbom, H.~Ahlstr{\"o}m, and
  J.~Kullberg, ``A concept for holistic whole body mri data analysis,
  imiomics,'' \emph{PloS one}, vol.~12, no.~2, p. e0169966, 2017.

\bibitem{christensen2001consistent}
G.~E. Christensen and H.~J. Johnson, ``Consistent image registration,''
  \emph{IEEE Transactions on Medical Imaging}, vol.~20, no.~7, pp. 568--582,
  2001.

\end{thebibliography}
\end{document}